\documentclass[runningheads]{llncs}
\usepackage[T1]{fontenc}
\usepackage{graphicx}
\usepackage{times}
\usepackage{epsfig}
\usepackage{amsmath}
\usepackage{amssymb} 
\usepackage{amsfonts}
\usepackage{bbm}
\usepackage{mathabx}
\usepackage{array}
\usepackage{xcolor}
\usepackage{tabu}
\usepackage{multirow}
\newcolumntype{M}[1]{>{\centering\arraybackslash}m{#1}}
\newcolumntype{L}[1]{>{\flushleft\arraybackslash}m{#1}}
\definecolor{mygray}{gray}{0.6}
\newcommand{\etal}{\textit{et al.}}

\begin{document}
%
\title{Surgical Triplet Recognition via Diffusion Model}

%
%
%
%
%

\author{
    Daochang Liu$^1$, 
    Axel Hu$^1$, 
    Mubarak Shah$^2$, 
    Chang Xu$^1$ \\
}

\authorrunning{D. Liu et al.}

\institute{ 
School of Computer Science, Faculty of Engineering, The University of Sydney\\ \and 
Center for Research in Computer Vision, University of Central Florida\\
{\tt\small \{daochang.liu, xintao.hu, c.xu\}@sydney.edu.au \quad shah@crcv.ucf.edu}\\
}

\maketitle              
\begin{abstract} 
Surgical triplet recognition is an essential building block to enable next-generation context-aware operating rooms. The goal is to identify the combinations of instruments, verbs, and targets presented in surgical video frames. In this paper, we propose \textit{DiffTriplet}, a new generative framework for surgical triplet recognition employing the diffusion model, which predicts surgical triplets via iterative denoising. To handle the challenge of triplet association, two unique designs are proposed in our diffusion framework, i.e., association learning and association guidance. During training, we optimize the model in the joint space of triplets and individual components to capture the dependencies among them. At inference, we integrate association constraints into each update of the iterative denoising process, which refines the triplet prediction using the information of individual components. Experiments on the CholecT45 and CholecT50 datasets show the superiority of the proposed method in achieving a new state-of-the-art performance for surgical triplet recognition. Our codes will be released.
\end{abstract}

\section{Introduction}

Surgical workflow analysis is a pivotal topic in surgical data science~\cite{maier2022surgical}, which facilitates intraoperative decision making and postoperative skill assessment~\cite{liu2021towards}. 
The concept of surgical action triplet~\cite{Tripnet} has emerged as an intriguing representation that provides fine-grained information about procedural activities and interactions within the surgical scene. 
The goal is to identify triplets of <\textit{instrument}, \textit{verb}, \textit{target}> by analyzing surgical videos captured by minimally invasive devices.
This task has been studied either as a recognition problem where each video frame is classified into categories of triplets~\cite{CholecTriplet2021}, or as a detection task which additionally pinpoints the spatial location of the triplets~\cite{CholecTriplet2022,sharma2023surgical}.
This paper focuses on the recognition of surgical action triplets.

Previous works on surgical triplet recognition have explored various models spanning from convolutional neural networks~\cite{Tripnet,cheng2023deep,ramesh2023dissecting}, recurrent neural networks~\cite{MTFiST}, graph neural networks~\cite{ConceptNet,ForestGCN,sista2022m}, to transformers~\cite{Rendezvous,SelfDistillSwin}.
Rendezvous~\cite{Rendezvous,RiT} is a milestone work that employs an instrument-centric attention mechanism, while Xi \etal ~\cite{ChainOfLook} take a verb-centric approach using verb prompts with vision-language models.
Researchers have also investigated on reducing the learning complexity of surgical triplets using progressive task disentanglement~\cite{TDN} or knowledge distillation techniques~\cite{MT4MTLKD,SelfDistillSwin}.
Unlike these previous works, this paper proposes a fundamentally new framework, \textit{DiffTriplet}, which tackles surgical triplet recognition in a generative learning way using the diffusion model.
The core idea is to learn to denoise noisy triplets in training and iteratively predict triplets by denoising from pure noise at inference.

Our diffusion-based framework is uniquely designed to address one of the key challenges regarding surgical triplets, i.e., the association of triplet components.
A surgical triplet comprises an instrument, a verb, and a target, where the association among them is an intrinsic property.
On one hand, the three individual components are interdependent, which means that the occurrence of one component may rely on other components. 
For example, the action verb <\textit{retract}> is more likely to be operated using the instrument <\textit{grasper}>, and the usual targets for this pair are <\textit{gallbladder}> and <\textit{liver}> if the specialty is cholecystectomy.
On the other hand, the triplet shows strong prior dependency on the individual components since the triplet classes, e.g. <\textit{grasper}, \textit{retract}, \textit{liver}>, are composited from the individual ones by definition.
The search space of the triplet will be greatly narrowed down if one of the individual components can be determined.
For instance, if the instrument is <\textit{grasper}>, the possible triplets will be only those related to this instrument.
Therefore, how to correctly build the association is essential for surgical triplet recognition.
To handle the challenge mentioned above, we propose two novel designs in our diffusion framework, which are \textit{association learning in training} and \textit{association guidance at inference}.
During the training process, our model learns to denoise in the joint space of triplets, instruments, verbs, and targets, so that the dependency relationships among triplets and individual components can be naturally captured by the generative learning capability of the diffusion model.
During the inference process, we propose an association guidance method to enforce the consistency between the predicted triplets and the predicted individual components, by incorporating the dependency constraints into the inference process of the diffusion model.
These two designs help our diffusion model better exploit the triplet association during both training and inference.

To validate our proposed method, we conduct comprehensive experiments on the CholecT45~\cite{CholecTriplet2021} and CholecT50~\cite{Rendezvous} datasets, where we achieve the state-of-the-art performance under the official cross-validation setting with 40.2\% average precision on CholecT45 and 40.3\% on CholecT50.
Experiments also show that the proposed association learning and association guidance bring evident improvements.
In summary, our contributions are three-fold:

\noindent
1) We propose a new generative framework for surgical triplet recognition which is the first diffusion model for surgical video understanding to our knowledge.

\noindent
2) We propose the joint space learning and the association guidance to explicitly enhance the correct triplet associations during both training and inference.

\noindent
3) We conduct extensive experiments to demonstrate our state-of-the-art performance.


\section{Method}

\newcommand{\mathxs}{\mathrm{x}}

The task of surgical triplet recognition is to take a surgical video as input and output the triplets in each frame.
A triplet is a combination of instrument, verb, and target.
There can be a single triplet, multiple triplets, or no triplet in each frame.
In the following, we begin with the background knowledge of diffusion models and a basic diffusion framework for triplet recognition.
We then introduce the designs of joint space learning and association guidance to handle the challenge of triplet association.

\begin{figure}
\includegraphics[width=\textwidth]{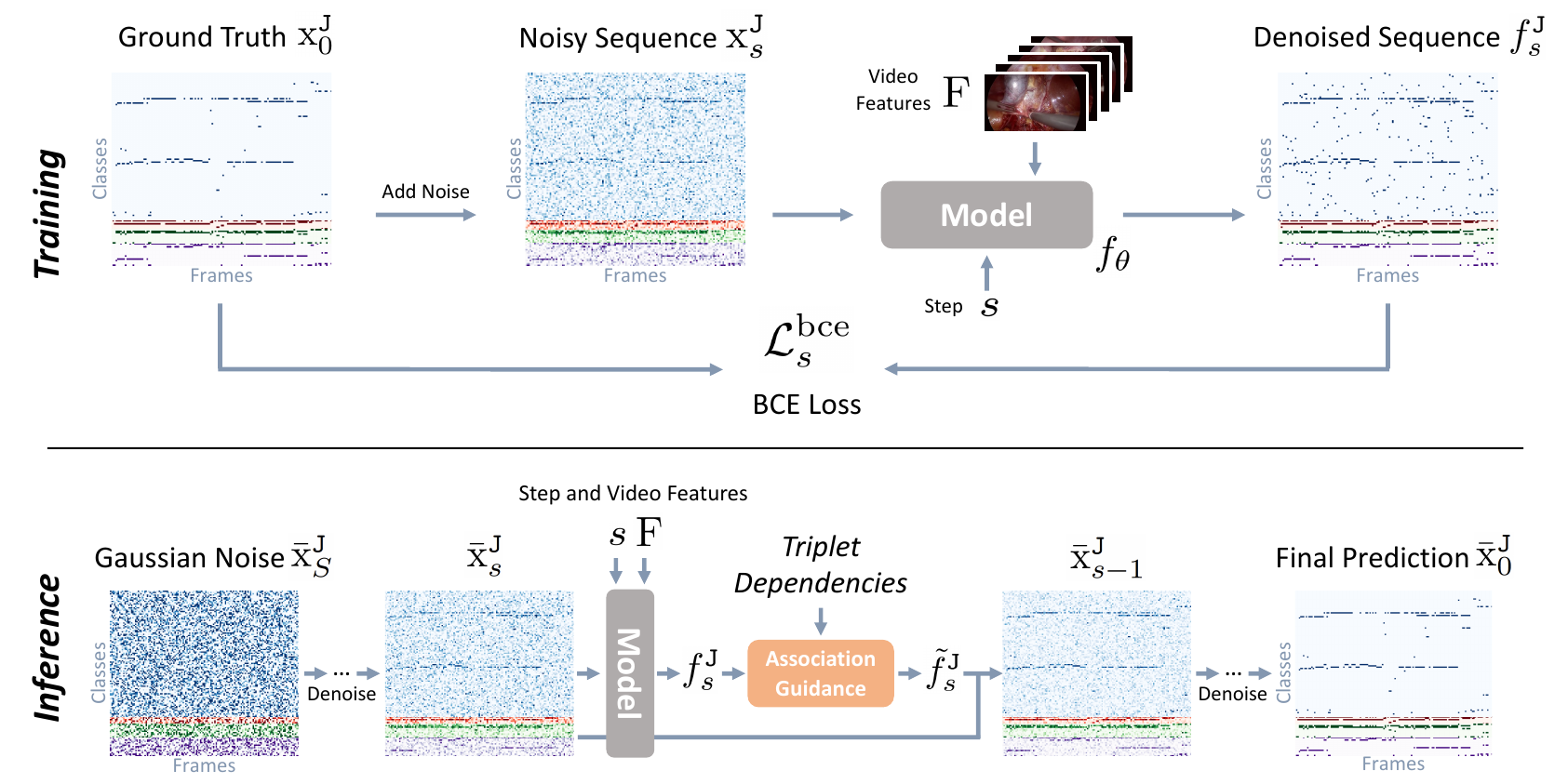}
\caption{Illustration of our \textit{DiffTriplet} framework. 
During training, the model is trained to denoise noisy sequences in the joint space of triplets and individual components. During inference, the model iteratively generates the prediction by gradually reducing the noise starting from a pure noise sequence. In the joint space matrices $\mathxs^{\mathtt{J}}, \bar{\mathxs}^{\mathtt{J}}$, the blue rows are triplets $\mathxs^{\mathtt{IVT}}, \bar{\mathxs}^{\mathtt{IVT}}$, red rows are instruments $\mathxs^{\mathtt{I}}, \bar{\mathxs}^{\mathtt{I}}$, green rows are verbs $\mathxs^{\mathtt{V}}, \bar{\mathxs}^{\mathtt{V}}$, and purple rows are targets $\mathxs^{\mathtt{T}}, \bar{\mathxs}^{\mathtt{T}}$. Darker colors means higher probabilities. For the model $f_\theta$, a causal temporal model is used.} \label{fig:overview}
\end{figure} 

\subsection{Preliminaries}
Diffusion models~\cite{BeatsGAN,DDPM,DDIM,LDM} have witnessed remarkable success as a new type of generative model, which have also shown potential for video analysis~\cite{DiffAct} in computer vision.
The diffusion model consists of two complementary processes, namely the forward process and the reverse process, to simulate the true distribution $q(\mathxs_0)$ with a model distribution $p_{\theta}(\mathxs_0)$.
Specifically, the forward process gradually adds noise into the clean data $\mathxs_0 \sim q(\mathxs_0)$ and produces a series of noisy data $\mathxs_1, \mathxs_2, ..., \mathxs_S$, where $S$ is the total steps of the process.
The reverse process generates new data starting from a pure Gaussian noise $\bar{\mathxs}_{S}$ and iteratively denoising the data along steps from $\bar{\mathxs}_{S-1}, \bar{\mathxs}_{S-2}, ... $ to $\bar{\mathxs}_0 \sim p_{\theta}(\mathxs_0)$ at the end. 



\subsection{Diffusion Triplet Recognition}
\label{sec:basic_framework}

Our diffusion framework for triplet recognition conducts generative learning on triplet sequences.
Given a surgical video with $L$ frames, we denote the ground-truth triplets as a binary sequence $\mathxs^{\mathtt{IVT}}_{0} \in \{0,1\}^{L \times C_{\mathtt{IVT}}}$, where $C_{\mathtt{IVT}}$ is the number of triplet classes.

\textbf{Training}. 
During training, the forward process gradually corrupts the ground-truth triplet sequence $\mathxs^{\mathtt{IVT}}_{0}$ to $\mathxs^{\mathtt{IVT}}_{1}, \mathxs^{\mathtt{IVT}}_{2}, ..., \mathxs^{\mathtt{IVT}}_{S}$ with varying noise levels.
Formally, at step $s$, the noisy triplet sequence $\mathxs^{\mathtt{IVT}}_s \in [0,1]^{L \times C_{\mathtt{IVT}}}$ is obtained in a closed form,
\begin{equation}\label{eq:forward}
\mathxs^{\mathtt{IVT}}_s = \sqrt{\alpha_s}\mathxs^{\mathtt{IVT}}_{0} + \epsilon \sqrt{1-\alpha_s},
\end{equation}
where $\epsilon$ is a random Gaussian noise and $\alpha_s$ is the schedule parameter defining the noise levels~\cite{DDIM}.
The resultant sequence is clipped within the range $[0,1]$.
We use a step-dependent neural network $f_\theta(\mathxs^{\mathtt{IVT}}_{s},s,\mathrm{F})$ to model the data distribution of triplet sequences by restoring the clean triplet sequence $\mathxs^{\mathtt{IVT}}_{0}$ from the noisy version $\mathxs^{\mathtt{IVT}}_s$.
The inputs of the network include the noisy sequence $\mathxs^{\mathtt{IVT}}_s$, the embedding of step $s$, and pre-extracted video features $\mathrm{F} \in \mathbb{R}^{L \times D}$ with $L$ frames and $D$ dimensions.
Using video features as conditional inputs for our diffusion model is crucial to build the mapping between surgical video and triplets in this video.
The network $f_\theta$ is then trained with a binary cross entropy (BCE) loss between the network output and the clean triplet sequence $\mathxs^{\mathtt{IVT}}_{0}$.
In each training iteration, the step $s \in \{1,2,...,S\}$ is selected randomly.
Note that we follow the convention of triplet recognition and use a \textit{causal} model for $f_\theta$ where the prediction of each frame can only utilize the information not after this frame.

\textbf{Inference.}
Once the network $f_\theta$ is optimized, the reverse process starts from pure Gaussian noise $\bar{\mathxs}^{\mathtt{IVT}}_{S}$ and iteratively reduces the noise through steps $\bar{\mathxs}^{\mathtt{IVT}}_{S-1}, \bar{\mathxs}^{\mathtt{IVT}}_{S-2}, ...$ to obtain the final prediction $\bar{\mathxs}^{\mathtt{IVT}}_0$ at the last step.
We use the bar notation $\bar{\cdot}$ to differentiate the variables during inference from those in training.   
At each inference step, the iteration adopts the update rule of the reverse sampling of diffusion models~\cite{DDIM},
\begin{equation}\label{eq:reverse}
\begin{split}
\bar{\mathxs}^{\mathtt{IVT}}_{s-1} = & \sqrt{\alpha_{s-1}} f_\theta(\bar{\mathxs}^{\mathtt{IVT}}_{s},s,\mathrm{F}) + \\
& \sqrt{1-\alpha_{s-1} - \sigma_s^2} \frac{\bar{\mathxs}^{\mathtt{IVT}}_{s}-\sqrt{\alpha_s}f_\theta(\bar{\mathxs}^{\mathtt{IVT}}_{s},s,\mathrm{F})}{\sqrt{1-\alpha_s}} + \sigma_s \epsilon,
\end{split}
\end{equation}
where $\epsilon$ is a random Gaussian noise, $\sigma_s$ is a variance schedule depending on the schedule parameter $\alpha_s$. 
The output $\bar{\mathxs}^{\mathtt{IVT}}_0$ in the last step is used as the final prediction since it is guaranteed to be close to the true triplets $\mathxs^{\mathtt{IVT}}_0$ according to diffusion model theories~\cite{DDPM}.

\subsection{Association Learning in Joint Space}

To better establish the association of triplet components, we first enhance the association learning by transforming our diffusion model from learning solely in the triplet space $\mathxs^{\mathtt{IVT}}_{0}$ to learning in a joint space $\mathxs^{\mathtt{J}}_{0}$ of both triplets and individual components.
Our framework is illustrated in Fig.~\ref{fig:overview}. 

\textbf{Joint space construction.} 
Recall that we conduct the forward and reverse processes on the triplet sequence $\mathxs^{\mathtt{IVT}}_{0}$ in Sec~\ref{sec:basic_framework}, and now we re-formulate the processes on a sequence $\mathxs^{\mathtt{J}}_{0} \in \{0,1\}^{L \times C_{\mathtt{J}}}$ in the joint space.
Specifically, the joint space $\mathxs^{\mathtt{J}}_{0}:=\mathxs^{\mathtt{IVT}}_{0} \oplus \mathxs^{\mathtt{I}}_{0} \oplus \mathxs^{\mathtt{V}}_{0} \oplus \mathxs^{\mathtt{T}}_{0}$ is constructed by combining the triplets with the corresponding instruments, verbs, and targets.
Let $\mathxs^{\mathtt{I}}_{0} \in \{0,1\}^{L \times C_{\mathtt{I}}}$, $\mathxs^{\mathtt{V}}_{0} \in \{0,1\}^{L \times C_{\mathtt{V}}}$, $\mathxs^{\mathtt{T}}_{0} \in \{0,1\}^{L \times C_{\mathtt{T}}}$ represent the binary ground-truth sequences for the instruments, verbs, and targets respectively, where $C_{\mathtt{I}}$, $C_{\mathtt{V}}$, $C_{\mathtt{T}}$ are the numbers of classes.
We choose $\oplus$ as a concatenation operation in our current implementation and thus the dimension of the joint space $C_{\mathtt{J}}$ adheres to $C_{\mathtt{J}} = C_{\mathtt{IVT}} + C_{\mathtt{I}} + C_{\mathtt{V}} + C_{\mathtt{T}}$.

\textbf{Joint space learning.} 
In the joint space, we reformulate the forward process in Eq.~\ref{eq:forward} to the following one to obtain the noisy sequence $\mathxs^{\mathtt{J}}_{s} \in [0,1]^{L \times C_{\mathtt{J}}}$ at step $s$,
\begin{equation}\label{eq:forward_joint}
\mathxs^{\mathtt{J}}_s = \sqrt{\alpha_s}\mathxs^{\mathtt{J}}_{0} + \epsilon \sqrt{1-\alpha_s}.
\end{equation}
We then replace the input of the network from the sequence in the triplet space $\mathxs^{\mathtt{IVT}}_{s}$ with the one in the joint space $\mathxs^{\mathtt{J}}_{s}$.
The network output $f_\theta(\mathxs^{\mathtt{J}}_{s},s,\mathrm{F})$ is denoted as $f^{\mathtt{J}}_{s} \in [0,1]^{L \times C_{\mathtt{J}}}$ for brevity.
We train the network to denoise the noisy sequence by minimizing a BCE loss between the output $f^{\mathtt{J}}_{s}$ and the ground-truth in the joint space $\mathxs^{\mathtt{J}}_{0}$,
\begin{equation}\label{eq:bce_loss}
\mathcal{L}^{\mathrm{bce}}_s = -\frac{1}{LC_{\mathtt{J}}}\sum_{i=1}^L\sum_{c=1}^{C_{\mathtt{J}}} \mathxs^{\mathtt{J}}_{0|i,c}\mathrm{log}f^{\mathtt{J}}_{s|i,c} + (1-\mathxs^{\mathtt{J}}_{0|i,c})\mathrm{log}(1-f^{\mathtt{J}}_{s|i,c}),
\end{equation}
where $\mathxs^{\mathtt{J}}_{0|i,c}$ and $f^{\mathtt{J}}_{s|i,c}$ are the ground truth and prediction elements for frame $i$ and class $c$ respectively.
The reverse process for inference in Eq.~\ref{eq:reverse} is updated correspondingly,
\begin{equation}\label{eq:reverse_joint}
\begin{split}
\bar{\mathxs}^{\mathtt{J}}_{s-1} = & \sqrt{\alpha_{s-1}} f_\theta(\bar{\mathxs}^{\mathtt{J}}_{s},s,\mathrm{F}) + \\
& \sqrt{1-\alpha_{s-1} - \sigma_s^2} \frac{\bar{\mathxs}^{\mathtt{J}}_{s}-\sqrt{\alpha_s}f_\theta(\bar{\mathxs}^{\mathtt{J}}_{s},s,\mathrm{F})}{\sqrt{1-\alpha_s}} + \sigma_s \epsilon.
\end{split}
\end{equation}
Such generative modeling using the diffusion model in the joint space naturally encourages the model to capture the dependencies among the triplets and the individual components during training.
Moreover, it also enables the injection of explicit triplet dependencies in the inference process with an association guidance we will introduce in the following section.


\subsection{Association Guidance at Inference}

Our second design to enhance the correct triplet association is a guidance method improving the inference process.
This is based on a key observation prevalent in existing triplet recognition models, where the models tend to have better recognition performance for individual components than the compositional triplet, potentially due to the higher dimensionality of triplet.
This phenomenon also exists in our diffusion model.
Therefore, we propose to utilize the individual component predictions to guide the triplet prediction and rectify errors by enforcing triplet association constraints.

To this end, we first represent the constraints with three dependency matrices $M_{\mathtt{I}} \in \{0,1\}^{C_{\mathtt{I}} \times C_{\mathtt{IVT}}}$, $M_{\mathtt{V}} \in \{0,1\}^{C_{\mathtt{V}} \times C_{\mathtt{IVT}}}$, $M_{\mathtt{T}} \in \{0,1\}^{C_{\mathtt{T}} \times C_{\mathtt{IVT}}}$ capturing the relations between triplets and instruments, verbs, targets respectively\footnote{Please refer to the supplementary material for visualized matrices.}.
The matrices register the possibility of triplets when each component occurs.
For example, $M_{\mathtt{I}}$ represents the possible triplets when each class of instrument appears.
These matrices can be obtained from the label mapping that defines the triplet classes or summarized from the statistics of the training dataset.
To inject the constraints, in each inference step, we decompose the model output in the joint space $f^{\mathtt{J}}_{s}$ back into individual spaces $f^{\mathtt{IVT}}_{s}, f^{\mathtt{I}}_{s}, f^{\mathtt{V}}_{s}, f^{\mathtt{T}}_{s}$,
\begin{equation}
f^{\mathtt{J}}_{s} \Rightarrow f^{\mathtt{IVT}}_{s} \in [0,1]^{L \times C_{\mathtt{IVT}}}, f^{\mathtt{I}}_{s} \in [0,1]^{L \times C_{\mathtt{I}}}, f^{\mathtt{V}}_{s} \in [0,1]^{L \times C_{\mathtt{V}}}, f^{\mathtt{T}}_{s} \in [0,1]^{L \times C_{\mathtt{T}}},
\end{equation}
where the decomposition a reversed operation of the previous concatenation.
Using the individual outputs, a guidance term $g_s \in [0,1]^{L \times C_{\mathtt{IVT}}}$ for the triplet prediction is computed incorporating the dependency between the triplet and the components,
\begin{equation}\label{eq:guidance}
g_s = (f^{\mathtt{I}}_{s} \otimes M_{\mathtt{I}}) \odot (f^{\mathtt{V}}_{s} \otimes M_{\mathtt{V}}) \odot (f^{\mathtt{T}}_{s} \otimes M_{\mathtt{T}}),
\end{equation}
where $\otimes$ is matrix multiplication and $\odot$ is element-wise product.
Intuitively, the guidance term in Eq.~\ref{eq:guidance} represents the possible triplets inferred from the predicted individual components, aggregated over instrument, verbs, and targets.
Then we compute a linear combination of the original triplet prediction $f^{\mathtt{IVT}}_{s}$ and the prediction combining the guidance term to steer the inference process with the triplet dependencies injected,
\begin{equation}
\tilde{f}^{\mathtt{IVT}}_{s} := (1-\omega)f^{\mathtt{IVT}}_{s} + \omega g_s \odot f^{\mathtt{IVT}}_{s}, 
\end{equation}
where $\omega$ is a hyper-parameter controlling the guidance scale.
The updated triplet prediction $\tilde{f}^{\mathtt{IVT}}_{s}$ alleviates the errors in $f^{\mathtt{IVT}}_{s}$ by employing the individual components that are usually better recognized.
The joint space is then re-composed using the updated triplet prediction, namely, $\tilde{f}^{\mathtt{J}}_{s}:=\tilde{f}^{\mathtt{IVT}}_{s} \oplus f^{\mathtt{I}}_{s} \oplus f^{\mathtt{V}}_{s} \oplus f^{\mathtt{T}}_{s}$.
The overall inference process is kept similar to Eq.~\ref{eq:reverse_joint} but only replacing $f^{\mathtt{J}}_{s}$ with the updated $\tilde{f}^{\mathtt{J}}_{s}$.
Our association guidance improves the inference process through the triplet association constraints.








\section{Experiments}


\textbf{Datasets.}
We conduct experiments on the benchmarks for surgical triplet recognition, i.e., CholecT50~\cite{Rendezvous} and CholecT45~\cite{CholecTriplet2021} datasets. 
The CholecT50 dataset consists of 50 videos of cholecystectomy surgeries captured by laparoscopic cameras, and the CholecT45 dataset is a subset with 45 videos.
The data has been used in the MICCAI EndoVis CholecTriplet challenges~\cite{CholecTriplet2021,CholecTriplet2022}.
Both datasets provide annotations for 100 classes of surgical triplets ($C_{\mathtt{IVT}}=100$), 6 classes of instruments ($C_{\mathtt{I}}=6$), 10 classes of verbs ($C_{\mathtt{V}}=10$), and 15 classes of targets ($C_{\mathtt{T}}=15$).
The videos are down-sampled to 1 fps and have 2000 frames per video on average.

\noindent
\textbf{Setup.}
Various experimental settings with different data splits have been employed in previous works~\cite{DataSplits}, making it challenging to compare across methods.
Therefore, we base our experiments on a consistent setup, using the five-fold \textbf{cross-validation setting recommended for research use} in the future, as described by the dataset owner in~\cite{DataSplits}.
We also use the official data splits~\cite{DataSplits}.
To evaluate recognition performance, the recommended Average Precision metrics~\cite{DataSplits} are computed for the triplet ($\mathrm{AP}_\mathtt{IVT}$), three individual components ($\mathrm{AP}_\mathtt{I}$, $\mathrm{AP}_\mathtt{V}$, $\mathrm{AP}_\mathtt{T}$), and the two instrument-related pairs ($\mathrm{AP}_\mathtt{IV}$, $\mathrm{AP}_\mathtt{IT}$). 
We use the official \texttt{ivtmetrics} library~\cite{DataSplits} to compute the metrics.

\noindent
\textbf{Implementation Details.}
The triplet sequences are normalized to $[-1, 1]$ when adding and removing noise.
Our model $f_\theta$ uses the architecture of the encoder in ASFormer~\cite{ASFormer}, which incorporates both temporal convolutions and attentions to enable temporal reasoning across frames.
We modify the model to be \textit{causal} to follow previous conventions regarding the causality constraint~\cite{CholecTriplet2021} that restricts the model from utilizing future frames.
This is achieved by changing the temporal convolutions to causal convolutions~\cite{TCN} and limiting the attention within the receptive fields of the causal convolutions. 
The model is of 2 layers with 256 feature maps.
We set the total number of diffusion step $S=1000$ and use the DDIM sampling rule~\cite{DDIM} with 8 steps at inference.
The model is trained using the Adam optimizer with a learning rate of 5e-5, a weight decay of 1e-5, a batch size of 1, and 500 epochs.
We set the guidance scale $\omega=1.0$ in our experiments.
The choice of the video features $\mathrm{F}$ is flexible.
In practice, given the code availability, we extract video features $\mathrm{F}$ from intermediate layer of pre-trained Rendezvous (RDV) models~\cite{Rendezvous} ($D=2500$) or pre-trained non-ensemble models of self-distilled-swin (SDSwin)~\cite{SelfDistillSwin} ($D=1024$).
The dependency matrices are obtained from the label mapping coming along with the dataset~\cite{Rendezvous}.
Our code is implemented using PyTorch.
The models are trained within one hour using a single NVIDIA V100 GPU.
Our full codes and pre-trained models will be released.

\begin{table}[t]
\begin{center}
\caption{Comparison to the state-of-the-art. We follow the official recommended cross-validation. $\dagger$: Results reported in original papers use the Rendezvous setting~\cite{Rendezvous} instead of cross-validation. $\ddagger$: Results reported in original papers use the Challenge setting~\cite{DataSplits} instead of cross-validation. $\mathsection$: Extra textual information and language models are used.
$*$: Results obtained using public codes. {\color{mygray}Gray}: Results should be compared with caution due to different settings. Empty cells mean results are not reported in original papers. $\mathrm{F}_{RDV}$: Rendezvous features. $\mathrm{F}_{SDSwin}$: SDSwin features.}
\label{table:sota-ap-cv}
\begin{tabu}{l c c c c c c}
\hline
Method & $\mathrm{AP}_\mathtt{I}$ & $\mathrm{AP}_\mathtt{V}$ & $\mathrm{AP}_\mathtt{T}$ & $\mathrm{AP}_\mathtt{IV}$ & $\mathrm{AP}_\mathtt{IT}$ & $\mathrm{AP}_\mathtt{IVT}$ \\
\hline
\multicolumn{7}{c}{{\textit{CholecT45}}}\\
\hline
Tripnet~\cite{Tripnet,ConceptNet} & 89.9±1.0 & 59.9±0.9 & 37.4±1.5 & 31.8±4.1 & 27.1±2.8 & 24.4±4.7 \\
Attention Triplet~\cite{Rendezvous,ConceptNet} & 89.1±2.1 & 61.2±0.6 & 40.3±1.2 & 33.0±2.9 & 29.4±1.2 & 27.2±2.7 \\
Rendezvous~\cite{Rendezvous,ConceptNet} & 89.3±2.1 & 62.0±1.3 & 40.0±1.4 & 34.0±3.3 & 30.8±2.1 & 29.4±2.8 \\
ConceptNet-ViT~\cite{ConceptNet} & 88.3±1.4 & 65.1±3.9 & 43.8±3.1 & 33.7±3.2 & 33.4±3.2 & 30.6±1.9 \\
RiT~\cite{RiT} & 88.6±2.6 & 64.0±2.5 & 43.4±1.4 & 38.3±3.5 & 36.9±1.0 & 29.7±2.6 \\
TDN~\cite{TDN} & 91.2±1.9 & 65.3±2.8 & 43.7±1.6 & - & - & 33.8±2.5 \\
MT4MTL-KD (No Ensemble)~\cite{MT4MTLKD}\;\; & 93.1±2.1 & \textbf{71.8}±3.4 & 48.8±3.8 & 44.9±2.4 & 43.1±2.0 & 37.1±0.5 \\
SDSwin (No Ensemble)~\cite{SelfDistillSwin} & - & - & - & - & - & 36.1±\;\;-\; \\
Ours ($\mathrm{F}_{RDV}$) & 91.8±1.8 & 66.3±1.4 & 45.5±1.9 & 42.7±4.6 & 42.2±1.9 & 35.0±3.2\\
Ours ($\mathrm{F}_{SDSwin}$) & \textbf{95.1}±1.6 & 71.1±1.3 & \textbf{51.7}±3.3 & \textbf{46.2}±3.9 & \textbf{47.4}±2.4 & \textbf{40.2}±1.9\\
\hline
\multicolumn{7}{c}{{\textit{CholecT50}}}\\
\hline
Rendezvous~\cite{Rendezvous,DataSplits} & 89.4±2.0 & 60.4±2.8 & 40.3±2.2 & 34.5±2.8 & 31.8±1.0 & 29.4±2.5\\
\rowfont{\color{mygray}}
$\dagger, \mathsection$ Forest GCN~\cite{ForestGCN} & 93.1±\;\;-\; & 60.1±\;\;-\; & 40.2±\;\;-\; & 36.2±\;\;-\; & 37.5±\;\;-\; & 36.7±\;\;-\; \\
\rowfont{\color{mygray}}
$\dagger, \mathsection$ Chain-of-Look~\cite{ChainOfLook} & 94.1±\;\;-\; & 62.5±\;\;-\; & 41.9±\;\;-\; & 41.7±\;\;-\; & 39.5±\;\;-\; & 38.2±\;\;-\; \\
\rowfont{\color{mygray}}
$\ddagger$ MoCoV2~\cite{ramesh2023dissecting} & - & - & - & - & - & 35.7±\;\;-\;  \\
\rowfont{\color{mygray}}
$\ddagger$ MT-FiST~\cite{MTFiST} & 82.1±\;\;-\;  & 51.5±\;\;-\;  & 45.5±\;\;-\;  & 37.1±\;\;-\;  & 43.1±\;\;-\;  & 35.8±\;\;-\;  \\
\rowfont{\color{mygray}}
$\ddagger$ MT4MTL-KD (No Ensemble)~\cite{MT4MTLKD}\;\; & 86.8±\;\;-\;  & 57.5±\;\;-\;  & 46.3±\;\;-\;  & 39.5±\;\;-\;  & 41.3±\;\;-\;  & 37.2±\;\;-\;  \\
\rowfont{\color{mygray}}
$\ddagger$ SDSwin (No Ensemble)~\cite{SelfDistillSwin} & - & - & - & - & - & 37.3±\;\;-\; \\
$*$ SDSwin (No Ensemble)~\cite{SelfDistillSwin} & - & - & - & - & - & 36.2±1.9 \\
Ours ($\mathrm{F}_{RDV}$) & 92.6±1.7 & 66.2±2.1 & 45.0±2.1 & 42.5±4.3 & 42.4±1.3 & 35.1±2.7 \\
Ours ($\mathrm{F}_{SDSwin}$) & \textbf{95.3}±1.4 & \textbf{70.9}±0.8 & \textbf{53.1}±2.6 & \textbf{46.4}±4.1 & \textbf{48.2}±1.8 & \textbf{40.3}±2.5\\
\hline
\end{tabu}
\end{center}
\end{table}


\begin{table}[t]
\begin{center}
\caption{Ablation studies on CholecT50 cross-validation using $\mathrm{F}_{RDV}$. We also provide a comparison between the performance of the (online) causal model and an (offline) acausal model.}
\label{table:ablation-main}
\begin{tabular}{l c c c c c c}
\hline
Method & $\mathrm{AP}_\mathtt{I}$ & $\mathrm{AP}_\mathtt{V}$ & $\mathrm{AP}_\mathtt{T}$ & $\mathrm{AP}_\mathtt{IV}$ & $\mathrm{AP}_\mathtt{IT}$ & $\mathrm{AP}_\mathtt{IVT}$ \\
\hline
\multicolumn{7}{c}{{\textit{\textbf{a}. Impact of Joint Space Learning}}}\\
\hline
Rendezvous Baseline~\cite{Rendezvous,DataSplits} \;\;\;\;\;\; & 89.4±2.0 & 60.4±2.8 & 40.3±2.2 & 34.5±2.8 & 31.8±1.0 & 29.4±2.5\\
Ours (w/o Joint Space) & - & - & - & 40.3±4.8 & 39.9±1.6 & 32.6±2.2 \\ 
Ours (with Joint Space) \;\;\;\;\;\;\;\; $ $ & 92.6±1.7 & 66.2±2.1 & 45.0±2.1 & 42.5±4.3 & 42.4±1.3 & 35.1±2.7 \\
\hline
\multicolumn{7}{c}{{\textit{\textbf{b}. Different Combinations of $\mathtt{I,V,T}$} for the Joint Space}}\\
\hline
Ours ($\mathtt{IVT}$) & - & - & - & 40.3±4.8 & 39.9±1.6 & 32.6±2.2 \\ 
Ours ($\mathtt{IVT} + \mathtt{I}$) & 92.5±1.6 & - & - & 41.7±4.5 & 40.2±1.3 & 33.0±2.2 \\ 
Ours ($\mathtt{IVT} + \mathtt{V}$) & - & 65.7±2.6 & -& 40.5±5.0 & 40.5±1.8 & 32.9±2.5 \\ 
Ours ($\mathtt{IVT} + \mathtt{T}$) & - & - & 44.1±2.2 & 40.6±4.7 & 39.5±1.6 & 32.3±2.0 \\ 
Ours ($\mathtt{IVT} + \mathtt{IV}$) & - & - & - & 41.0±4.4 & 40.2±1.6 & 32.9±2.3 \\ 
Ours ($\mathtt{IVT} + \mathtt{IT}$) & - & - & - & 41.0±5.1 & 40.4±1.5 & 32.8±2.3 \\ 
Ours ($\mathtt{IVT} + \mathtt{I}  + \mathtt{V}  + \mathtt{T}$) & 92.6±1.7 & 66.2±2.1 & 45.0±2.1 & 42.5±4.3 & 42.4±1.3 & 35.1±2.7 \\
\hline
\multicolumn{7}{c}{{\textit{\textbf{c}. Impact of Association Guidance}}}\\
\hline
Ours (Scale $\omega = 0.00)$ & 92.6±1.7 & 66.2±2.1 & 45.0±2.1 & 41.2±4.3 & 40.9±1.4 & 33.3±2.2 \\
Ours (Scale $\omega = 0.50)$ & 92.6±1.7 & 66.2±2.1 & 45.0±2.1 & 41.7±4.5 & 41.4±1.4 & 33.9±2.3 \\
Ours (Scale $\omega = 1.00)$ & 92.6±1.7 & 66.2±2.1 & 45.0±2.1 & 42.5±4.3 & 42.4±1.3 & 35.1±2.7 \\
\hline
\multicolumn{7}{c}{{\textit{\textbf{d}. Causal Model and Acausal Model}}}\\
\hline
Ours (Causal) & 92.6±1.7 & 66.2±2.1 & 45.0±2.1 & 42.5±4.3 & 42.4±1.3 & 35.1±2.7\\
Ours (Acausal) & 94.0±1.1 & 68.2±2.2 & 46.3±3.0 & 44.9±4.8 & 44.7±2.1 & 37.2±3.7\\
\hline
\end{tabular}
\end{center}
\end{table}

\subsection{Results}

\textbf{Comparison to the State-of-the-Art.}
Table~\ref{table:sota-ap-cv} presents a comparison between our results and previous results reported in the literature on the two datasets.
For previous methods using ensemble models~\cite{SelfDistillSwin,MT4MTLKD}, we present the results of their non-ensemble models to facilitate direct comparison.
It can be seen that our proposed method evidently improves the state-of-the-art performance (40.2 vs. 37.1 $\mathrm{AP}_\mathtt{IVT}$ on CholecT45), validating the efficacy of our diffusion-based framework for triplet recognition.
In Table~\ref{table:sota-ap-cv}, we also include results previously reported under the Rendezvous setting~\cite{Rendezvous} and the Challenge setting~\cite{DataSplits}, not officially recommended for research~\cite{DataSplits}, for readers' reference but with caution.
Our method only has a slight computational overhead, with an inference speed of 16.9 fps, compared to 19.1 fps for Rendezvous~\cite{Rendezvous}, both measured on the same device with a NVIDIA T4 GPU.




\textbf{Ablation Studies.}
We present ablation studies in Table~\ref{table:ablation-main} to validate each design.

\textit{Joint space learning.}
Firstly, the importance of the proposed joint space learning is investigated in Table~\ref{table:ablation-main}a.
Compared to the baseline, our diffusion model achieves notable gain (32.6 vs. 29.4 $\mathrm{AP}_\mathtt{IVT}$), which is further improved by learning in the joint space (35.1 vs. 32.6 $\mathrm{AP}_\mathtt{IVT}$).
We further provide the results using different combinations regarding the joint space in Table~\ref{table:ablation-main}b. 
Other combinations ($\mathtt{IVT} + \mathtt{I/V/T/IV/IT}$) are not as good as our proposed space ($\mathtt{IVT} + \mathtt{I} + \mathtt{V} + \mathtt{T}$), due to the partial association.

\textit{Association Guidance.}
Secondly, we provides the results with difference scales of the association guidance in Table~\ref{table:ablation-main}c, where $\omega = 1.00$ means full guidance and $\omega = 0.00$ means no guidance.
The guidance effectively strengthens the triplet association and obtains better performance (35.1 vs. 33.3 $\mathrm{AP}_\mathtt{IVT}$).
Metrics for individual components are the same across scales since they are not affected by the guidance.

\textit{Causal vs. Acausal.}
Thirdly, in Table~\ref{table:ablation-main}d, we additionally explore how the model performs when the causality constraint is removed. 
We compare the results of the causal model and an acausal model that employing the normal version of convolutions and attentions.
It is encouraging that the performance is further improved when the acausal model is used (37.2 vs. 35.1 $\mathrm{AP}_\mathtt{IVT}$), showing the potential of our method not only for online analysis but also more accurate offline analysis for surgical videos.

\textbf{Inference Steps.}
In Table~\ref{table:ablation-step}, we present the results using different numbers of inference steps. This experiment shows that the result is stable across different inference steps.
The computational cost increases linearly with the number of inference steps. Therefore, we use 8 steps to achieve a balance between accuracy and speed.

\begin{table}
\begin{center}
\caption{Results using different inference steps on CholecT50 cross-validation with $\mathrm{F}_{RDV}$.}
\label{table:ablation-step}
\begin{tabular}{l c c c c c c}
\hline
Steps & $\mathrm{AP}_\mathtt{I}$ & $\mathrm{AP}_\mathtt{V}$ & $\mathrm{AP}_\mathtt{T}$ & $\mathrm{AP}_\mathtt{IV}$ & $\mathrm{AP}_\mathtt{IT}$ & $\mathrm{AP}_\mathtt{IVT}$ \\
\hline
Steps = 1 & 91.9±1.7 & 65.4±2.3 & 44.6±2.2 & 41.9±4.4 & 41.5±1.7 & 34.3±3.0 \\
Steps = 2 & 92.5±1.7 & 65.9±2.3 & 44.9±2.2 & 42.4±4.4 & 42.3±1.4 & 35.0±2.9 \\
Steps = 4 & 92.6±1.7 & 66.1±2.1 & 45.1±2.1 & 42.6±4.3 & 42.4±1.4 & 35.1±2.7 \\
\underline{Steps = 8} & 92.6±1.7 & 66.2±2.1 & 45.0±2.1 & 42.5±4.3 & 42.4±1.3 & 35.1±2.7 \\
Steps = 16 & 92.6±1.7 & 66.1±2.1 & 45.0±2.1 & 42.5±4.3 & 42.4±1.3 & 35.0±2.7 \\
\hline
\end{tabular}
\end{center}
\end{table}

\section{Conclusion}

This paper proposes a generative framework for surgical triplet recognition using a diffusion model with the joint space learning and the association guidance.
Future works can explore more forms of the triplet dependencies, such as the soft dependencies summarized from training statistics, and the inter-component dependencies to refine the predictions of the individual components.
It is also promising to design an association guidance with a progressive hierarchy from components to pairs and finally to triplets.

\textbf{Acknowledgement}. The authors acknowledge the use of the National Computational Infrastructure (NCI) which is supported by the Australian Government and accessed through the NCI AI Flagship Scheme and Sydney Informatics Hub HPC Allocation Scheme.

\bibliographystyle{splncs04}
\bibliography{mybib}

\newpage

\section{Supplementary Material}

\begin{figure}
\includegraphics[width=\textwidth]{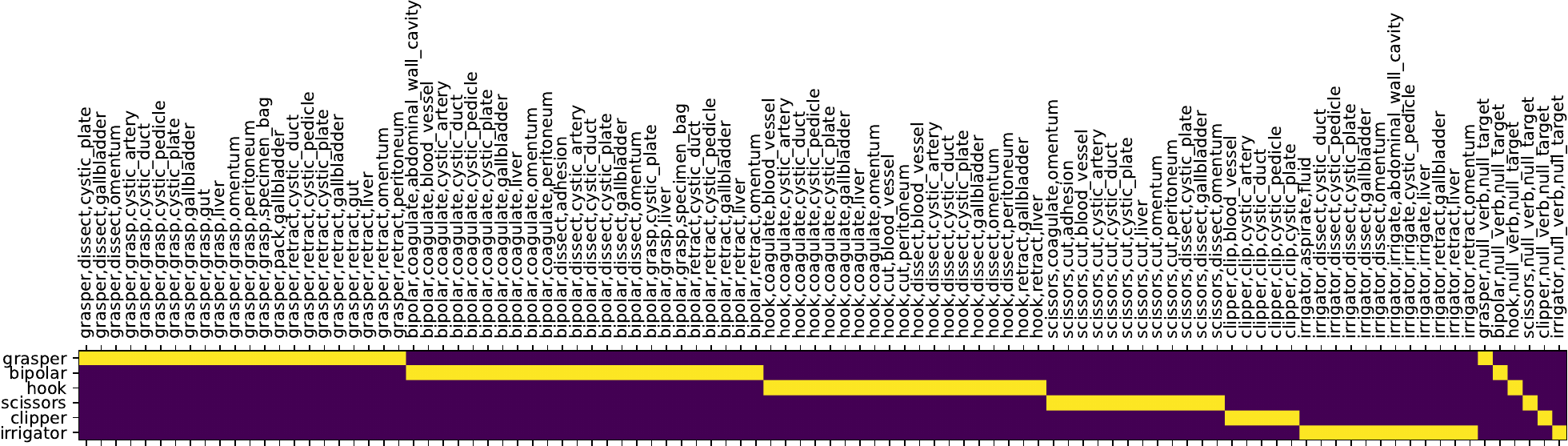}
\caption{Visualization of the dependency matrix $M_{\mathtt{I}} \in \{0,1\}^{C_{\mathtt{I}} \times C_{\mathtt{IVT}}}$. Yellow cells (value=1) are the possible associations between the triplet and the instrument. Blue cells (value=0) are the impossible associations between the triplet and the instrument.} \label{fig:supp_M_i}
\end{figure}

\begin{figure}
\includegraphics[width=\textwidth]{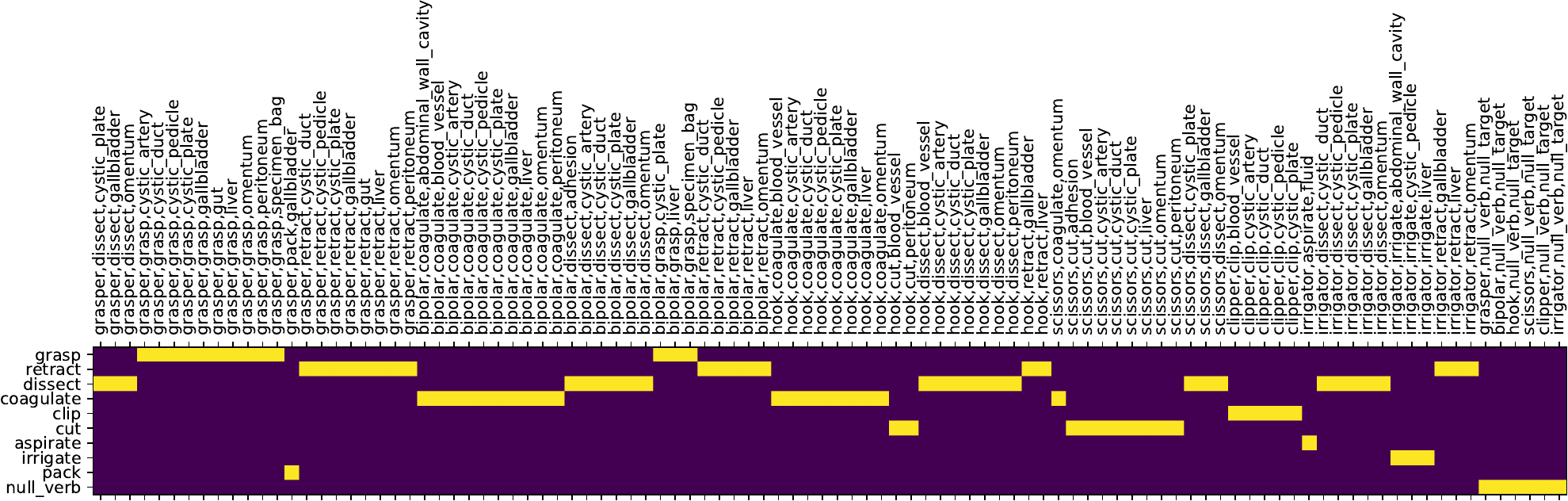}
\caption{Visualization of the dependency matrix $M_{\mathtt{V}} \in \{0,1\}^{C_{\mathtt{V}} \times C_{\mathtt{IVT}}}$. Yellow cells (value=1) are the possible associations between the triplet and the verb. Blue cells (value=0) are the impossible associations between the triplet and the verb.} \label{fig:supp_M_v}
\end{figure}

\begin{figure}
\includegraphics[width=\textwidth]{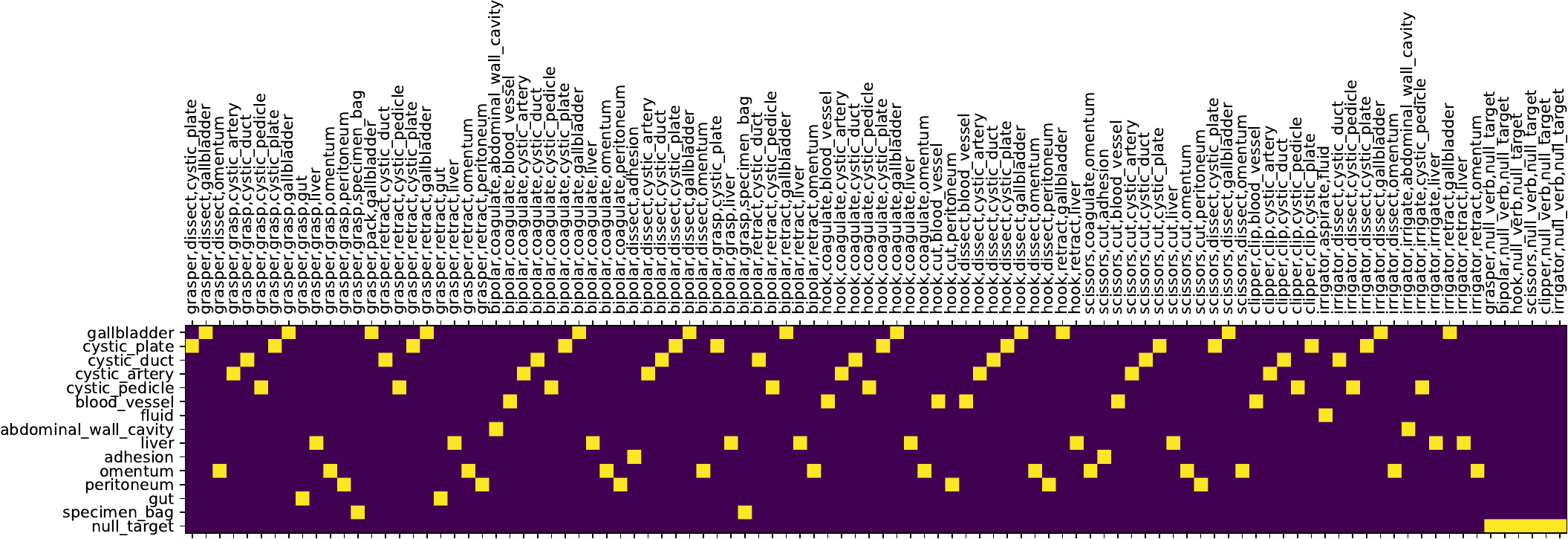}
\caption{Visualization of the dependency matrix $M_{\mathtt{T}} \in \{0,1\}^{C_{\mathtt{T}} \times C_{\mathtt{IVT}}}$. Yellow cells (value=1) are the possible associations between the triplet and the target. Blue cells (value=0) are the impossible associations between the triplet and the target.} \label{fig:supp_M_t}
\end{figure}

\end{document}